\documentclass[final]{cvpr}

\usepackage{times}
\usepackage{epsfig}
\usepackage{graphicx}
\usepackage{amsmath}
\usepackage{amssymb}

\usepackage{bm}
\usepackage{booktabs}
\usepackage{siunitx}
\usepackage{stfloats}
\usepackage{multirow}
\usepackage{pifont}

\newcommand{\xmark}{\ding{86}}

\usepackage[pagebackref=true,breaklinks=true,colorlinks,bookmarks=false]{hyperref}

\def\cvprPaperID{****} 
\def\confYear{CVPR ****}

\begin{document}

\title{MicroNet: Towards Image Recognition with Extremely Low FLOPs}

\author{{Yunsheng Li$^1$ \qquad Yinpeng Chen$^2$ \qquad Xiyang Dai$^2$ \qquad Dongdong Chen$^2$ \qquad Mengchen Liu$^2$}\\
{\qquad Lu Yuan$^2$ \qquad Zicheng Liu$^2$ \qquad Lei Zhang$^2$ \qquad Nuno Vasconcelos$^1$}\\
\\
$^1$UC San Diego \qquad $^2$Microsoft\\
{\tt\small yul554@eng.ucsd.edu, nvasconcelos@ucsd.edu}\\
{\tt\small \{yiche,xidai,dochen,mengcliu,luyuan,zliu,leizhang\}@microsoft.com}
}

\maketitle

\begin{abstract}
   In this paper, we present MicroNet, which is an efficient convolutional neural network using extremely low computational cost (e.g. 6 MFLOPs on ImageNet classification). Such a low cost network is highly desired on edge devices, yet usually suffers from a significant performance degradation. We handle the extremely low FLOPs based upon two design principles: (a) avoiding the reduction of network width by lowering the node connectivity, and (b) compensating for the reduction of network depth by introducing more complex non-linearity per layer. Firstly, we propose Micro-Factorized convolution to factorize both pointwise and depthwise convolutions into low rank matrices for a good tradeoff between the number of channels and input/output connectivity. Secondly, we propose a new activation function, named Dynamic Shift-Max, to improve the non-linearity via maxing out multiple dynamic fusions between an input feature map and its circular channel shift. The fusions are dynamic as their parameters are adapted to the input. Building upon Micro-Factorized convolution and dynamic Shift-Max, a family of MicroNets achieve a significant performance gain over the state-of-the-art in the low FLOP regime. For instance, MicroNet-M1 achieves 61.1\% top-1 accuracy on ImageNet classification with 12 MFLOPs, outperforming MobileNetV3 by 11.3\%.
\end{abstract}

\section{Introduction}
Recently, designing efficient CNN architectures \cite{squeezenet16,howard2017mobilenets,sandler2018mobilenetv2,Howard_2019_ICCV_mbnetv3,Zhang_2018_CVPR,ma_2018_ECCV,tan-ICML19-efficientnet} has been an active research area. These works enable high quality services on edge devices. However, even the state-of-the-art efficient CNNs (e.g. MobileNetV3 \cite{Howard_2019_ICCV_mbnetv3}) suffer from a significant performance degradation when the computational cost becomes extremely low. For instance, when constraining MobileNetV3 from 112M to 12M MAdds on image classification \cite{deng2009imagenet} at resolution 224 $\times$ 224, the top-1 accuracy drops  from 71.7\% to 49.8\%. This makes its adoption harder on low power devices (e.g. IoT devices). In this paper, we target a more challenging problem by cutting half of the budget: \textit{can we perform image classification over 1,000 classes at resolution $224 \times 224$ under 6 MFLOPs?}

This extremely low computational cost (6M FLOPs) requires a careful redesign of every layer. For instance, even a thin stem layer that contains a single $3 \times 3$ convolution with 3 input channels and 8 output channels over a $112 \times 112$ grid (stride=2) requires 2.7M MAdds. The resource left for designing convolution layers and a classifier for 1,000 classes is too limited to learn a good representation. To fit in such a low budget, a naive strategy for applying existing efficient CNNs (e.g. MobileNet \cite{howard2017mobilenets, sandler2018mobilenetv2, Howard_2019_ICCV_mbnetv3} and ShuffleNet \cite{Zhang_2018_CVPR, ma_2018_ECCV}) is to significantly reduce the width or depth of the network. This results in a severe performance degradation.

\begin{figure}[t]
	\begin{center}
		\includegraphics[width=0.8\linewidth]{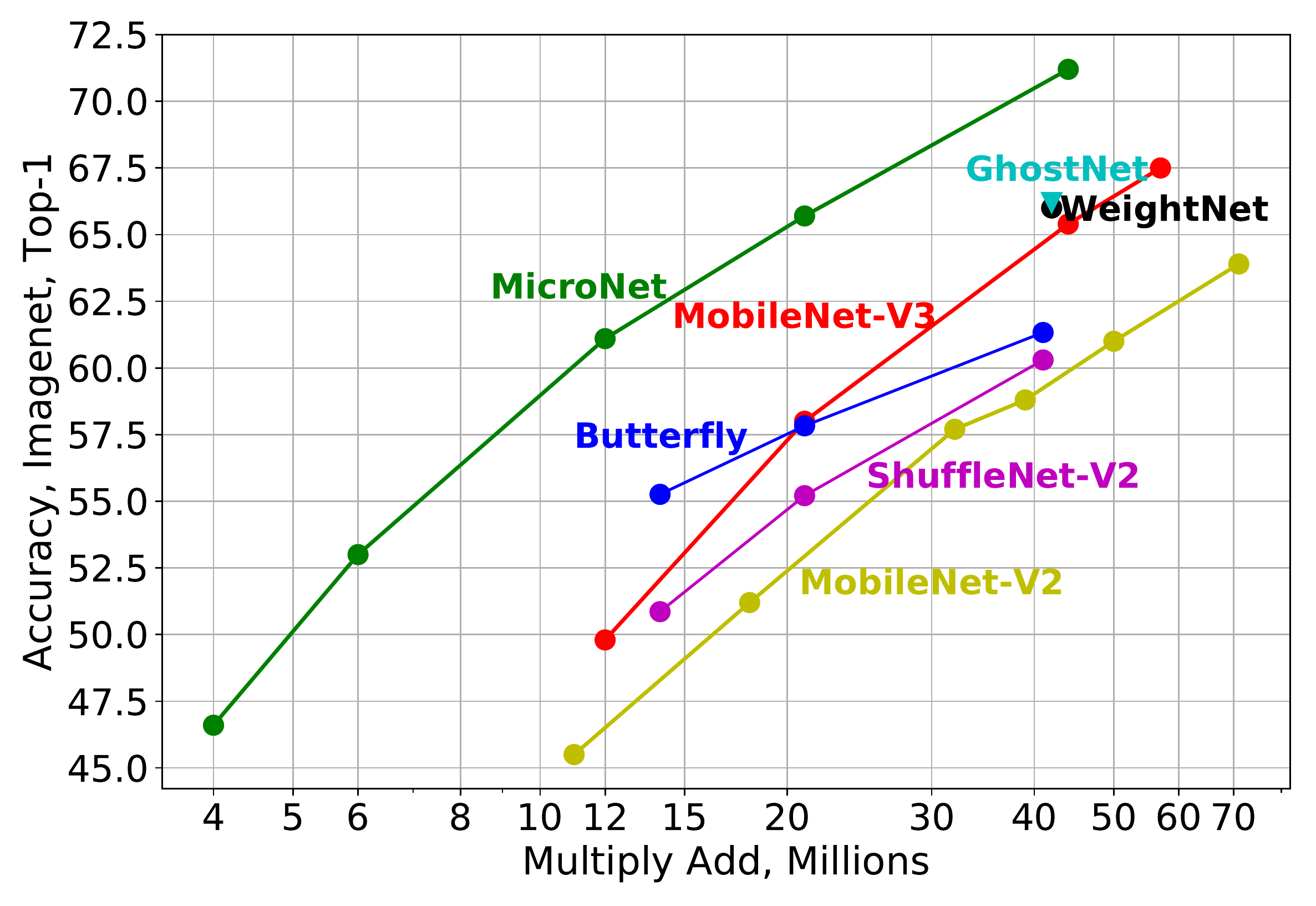}
	\end{center}
	\vspace{-1mm}
	\caption{\textbf{Computational Cost (MAdds) vs. ImageNet Accuracy.} MicroNet significantly outperforms the state-of-the-art efficient networks at very low FLOPs (from 4M to 45M MAdds). Best viewed in color.}
	\label{fig:micronet-teasor}
	\vspace{-1mm}
\end{figure}

We propose a new architecture, named MicroNet, to handle the extremely low FLOPs. It builds upon two design principles as follows: 
\begin{itemize}
    \vspace{-1mm} \item \textit{Circumventing the reduction of network width through lowering node connectivity.}
    \vspace{-1mm} \item \textit{Compensating for the reduction of network depth by improving non-linearity per layer.}
\end{itemize}
These principles guide us to design more efficient convolution and activation functions.

Firstly, we propose \textit{Micro-Factorized Convolution} to factorize both pointwise and depthwise convolutions into low rank matrices. This provides a good balance between the input/output connectivity and the number of channels. Specifically, 
we design \textit{group-adaptive convolution} to factorize pointwise convolution. It adapts the number of groups to the number of channels by a square root relationship. Stacking two group-adaptive convolutions essentially approximates a pointwise convolution matrix by a block matrix, of which each block has rank-1.
The factorization (rank-1) of depthwise convolution is straightforward, by factorizing a $k \times k$ depthwise convolution into a $1 \times k$ and a $k \times 1$ depthwise convolution. We show that a proper combination of these two approximations over different levels significantly reduces the computational cost without sacrificing the number of channels.

Secondly, we propose a new activation function, named \textit{Dynamic Shift-Max} to improve the non-linearity from two aspects: (a) it maxes out multiple fusions between an input feature map and its circular channel shift,
and (b) each fusion is dynamic as its parameters adapt to the input. Furthermore, it enhances both node connectivity and non-linearity efficiently in one function with a low computational cost.

Experimental results show that MicroNet outperforms the state-of-the-art by a large margin (see Figure \ref{fig:micronet-teasor}). For example, compared to MobileNetV3, our method gains 11.3\% and 7.7\% in top-1 accuracy on ImageNet classification, under the constraints of 12M and 21M FLOPs respectively. Within the extremely challenging 6 MFLOPs constraint, our method achieves 53.0\% top-1 accuracy, gaining 3.2\% over MobileNetV3 that has a doubled complexity (12 MFLOPs). In addition, a family of MicroNets provide strong baselines for two pixel-level tasks with very low computational cost: semantic segmentation and keypoint detection.

\section{Related Work}

\noindent \textbf{Efficient CNNs:}
MobileNets \cite{howard2017mobilenets, sandler2018mobilenetv2, Howard_2019_ICCV_mbnetv3} decompose $k \times k$ convolution into a depthwise and a pointwise convolution. ShuffleNets \cite{Zhang_2018_CVPR, ma_2018_ECCV} use group convolution and channel shuffle to simplify pointwise convolution. \cite{vahid_2020_CVPR} uses butterfly transform to approximate pointwise convolution. EfficientNet \cite{tan-ICML19-efficientnet, Tan_2020_CVPR} finds a proper relationship between input resolution and network width/depth. MixNet \cite{Tan-bmvc2019-mixconv} mixes up multiple kernel sizes in a single convolution. AdderNet \cite{Chen_2020_CVPR_addernet} trades massive multiplications for much cheaper additions. GhostNet \cite{Han_2020_CVPR_ghostnet} applies cheap linear transformations to generate ghost feature maps. Sandglass \cite{Daquan_2020_ECCV_RethinkingBS} flips the structure of inverted residual block to alleviate information loss.
\cite{yu2018slimmable} and \cite{Cai2019OnceFA} train one network to support multiple sub-networks. 

\noindent \textbf{Efficient Inference:} Efficient inference \cite{NIPS2017_6813,liu2018ddnn,Wang_2018_ECCV,Wu_2018_CVPR} customizes a proper sub-network adaptively per input. \cite{Wang_2018_ECCV} and \cite{Wu_2018_CVPR} use reinforcement learning to learn a controller for skipping part of an existing model. MSDNet \cite{huang2018multiscale} allows early-exit for easy samples based on the prediction confidence. \cite{Yuan2019S2DNASTS} searches for the optimal MSDNet. \cite{Yang_2020_CVPR} adapts image resolution to achieve efficient inference.

\noindent \textbf{Dynamic Neural Networks:}
Dynamic networks boost the representation capability by adapting parameters to the input. HyperNet \cite{Ha2017HyperNetworks} uses another network to generate parameters for the main network. SENet \cite{Hu_2018_CVPR} reweighs channels by squeezing global context. SKNet \cite{Li_2019_CVPR_SKNet} adapts attention over kernels with different sizes. Dynamic convolution \cite{Yang2019CondConvCP,Chen2019DynamicCA} aggregates multiple convolution kernels based on their attention. Dynamic ReLU \cite{Chen2020DynamicReLU} adapts slopes and intercepts of two linear functions in ReLU \cite{NairH10Relu,JarrettKRL09Relu}. 
\cite{Ma_2020_eccv_WeightNetRT} uses grouped fully connected layer to generate convolutional weights directly.
\cite{Chen2020DynamicRC} extends dynamic convolution from spatial agnostic to spatial specific. \cite{Su_2020_eccv_DynamicGC} proposes dynamic group convolution that adaptively groups input channels. \cite{Tian_2020_eccv_ConditionalCF} applies dynamic convolution on instance segmentation.  
\cite{Li_2020_CVPR_dynamic_routing} learns dynamic routing across scales for semantic segmentation.

\section{Our Method: MicroNet}
Below we describe in detail the design principles and key components of MicroNet.

\subsection{Design Principles}
The extremely low FLOPs constrain both the network width (number of channels) and network depth (number of layers), which are analyzed separately. If we consider a convolution layer as a graph, the connections (edges) between input and output (nodes) are weighted by kernel parameters. Here, we define the \textit{connectivity} as the number of connections per output node. Thus, the number of connections equals to the product of the number of output channels and the connectivity. When the computation cost (proportional to the number of connections) is fixed, the number of channels conflicts with the connectivity. We believe that a good balance between them can effectively avoid channel reduction and improve the representation capability of a layer. Therefore, our first design principle is: \textbf{\textit{circumventing the reduction of network width through lowering node connectivity.}} We achieve this by factorizing both pointwise and depthwise convolutions at a finer scale. 

When the depth (number of layers) is significantly reduced for a network, its non-linearity (encoded in ReLU) is constrained, resulting in a clear performance degradation. This motivates our second design principle as: \textbf{\textit{compensating for the reduction of network depth by improving non-linearity per layer.}} We achieve this by designing a new activation function, \textit{Dynamic Shift-Max}. 

\subsection{Micro-Factorized Convolution}
We factorize both pointwise and depthwise convolutions at a finer scale, from where Micro-Factorized convolution gets its name. The goal is to balance between the number of channels and input/output connectivity.

\vspace{1mm}
\noindent \textbf{Micro-Factorized Pointwise Convolution:}
\begin{figure*}[t]
	\begin{center}
		\includegraphics[width=1.0\linewidth]{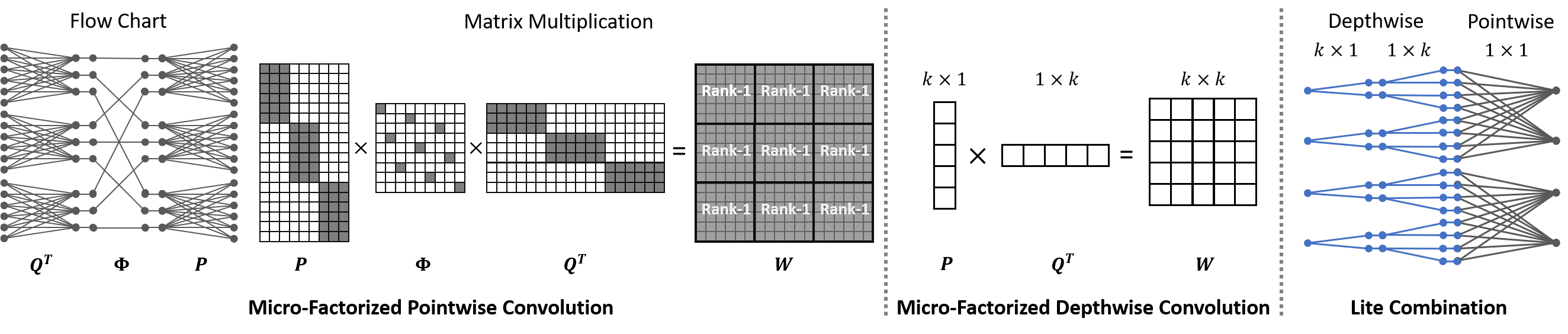}
	\end{center}
	\vspace{-3mm}
	\caption{\textbf{Micro-Factorized pointwise and depthwise convolutions}. \textbf{Left:} factorizing a pointwise convolution into two group-adaptive convolutions, where the group number $G=\sqrt{C/R}=\sqrt{18/2}=3$. The resulting matrix $\bm{W}$ can be divided into $G \times G$ blocks, of which each block has rank 1. \textbf{Middle:} factorizing a $k \times k$ depthwise convolution into a $k \times 1$ and a $1 \times k$ depthwise convolutions. \textbf{Right:} lite combination of Micro-Factorized pointwise and depthwise convolutions.}
	\label{fig:low-rank}
	\vspace{-2mm}
\end{figure*}
We propose group-adaptive convolution to factorize a pointwise convolution. For the sake of conciseness, we assume the convolution kernel $\bm{W}$ has the same number of input and output channels ($C_{in} = C_{out} = C$) and ignore the bias. The kernel matrix $\bm{W}$ is factorized into two group-adaptive convolutions, of which the group number $G$ depends on the number of channels $C$. Mathematically, it can be represented as:
\begin{align}
\bm{W} = \bm{P}\bm{\Phi}\bm{Q}^T,
\label{eq:mat-fac}
\end{align}
where $\bm{W}$ is a $C \times C$ matrix. $\bm{Q}$ is of shape $C \times \frac{C}{R}$, squeezing the number of channels by ratio $R$. $\bm{P}$ is of shape $C \times \frac{C}{R}$, expanding the number of channels back to $C$ as output. 
$\bm{P}$ and $\bm{Q}$ are diagonal block matrices with $G$ blocks, of which each block corresponds to the convolution of a group. $\bm{\Phi}$ is a $\frac{C}{R} \times \frac{C}{R}$ permutation matrix, shuffling channels similarly as in \cite{Zhang_2018_CVPR}. The computational complexity is $\mathcal{O}=\frac{2C^2}{RG}$.
Figure \ref{fig:low-rank}-Left shows an example with $C=18$, $R=2$ and $G=3$.

\begin{figure}[t]
	\begin{center}
		\includegraphics[width=0.7\linewidth]{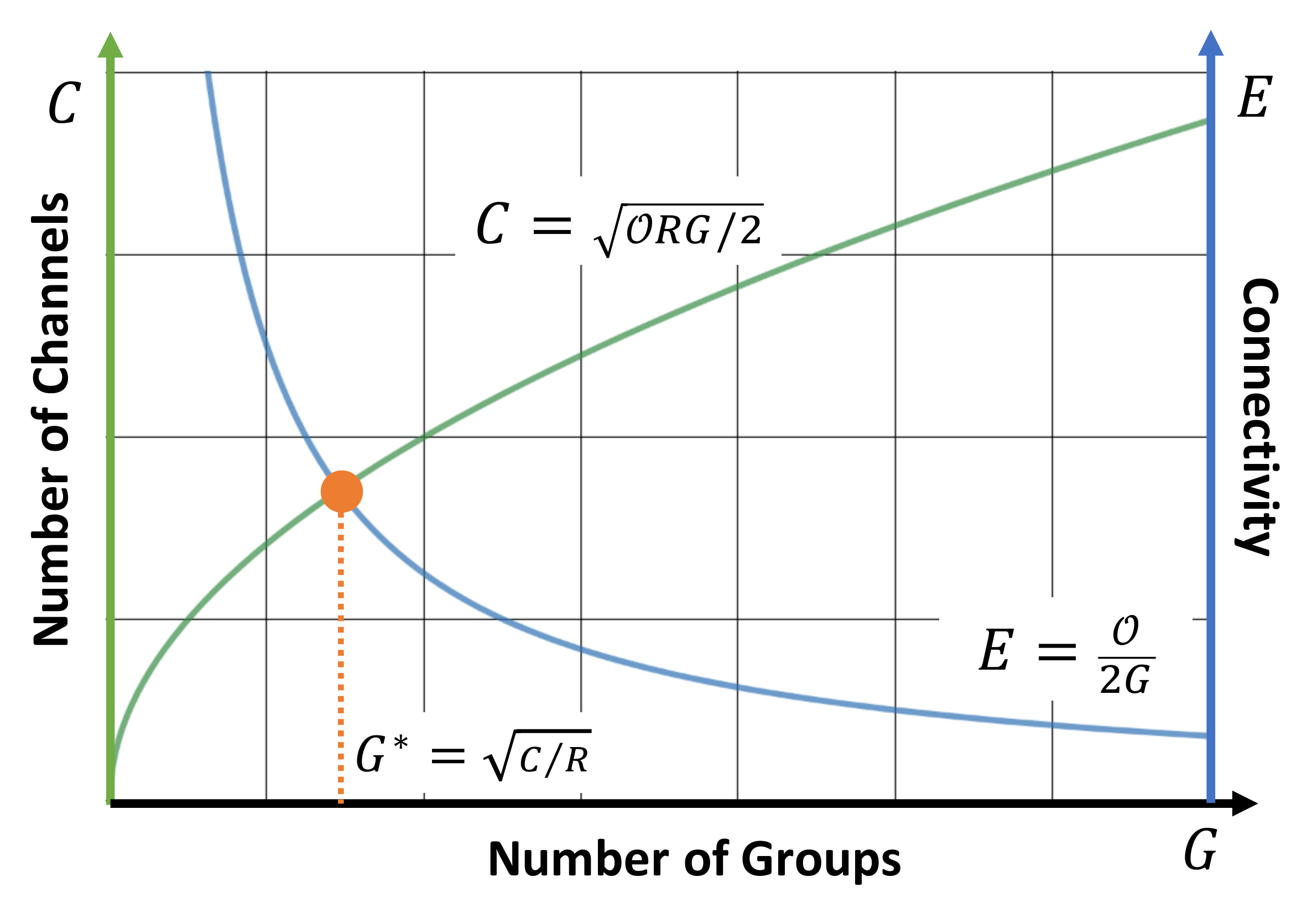}
	\end{center}
	\vspace{-2mm}
	\caption{\textbf{Number of Channels $C$ vs. Connectivity $E$} over number of groups $G$. We assume that the computational cost $\mathcal{O}$ and the reduction ratio $R$ are fixed. Best viewed in color.}
	\label{fig:hg2}
	\vspace{-2mm}
\end{figure}

Note that the group number $G$ is \textit{not} fixed but adapts to the number of channels $C$ and reduction ratio $R$ as:
\begin{align}
G=\sqrt{C/R}.
\label{eq:g-cr}
\end{align}
This square root relation is derived from balancing between the number of channels $C$ and input/output connectivity. Here, we define the \textit{connectivity} $E$ as the number of input-output connections per output channel. 
Each output channel connects to $\frac{C}{RG}$ hidden channels between the two group-adaptive convolutions, and each hidden channel connects to $\frac{C}{G}$ input channels. Thus $E=\frac{C^2}{RG^2}$.
When we fix the computational complexity $\mathcal{O}=\frac{2C^2}{RG}$ and the reduction ratio $R$, the number of channels $C$ and the connectivity $E$ change in opposite directions over $G$ as:
\begin{align}
C=\sqrt{\frac{\mathcal{O}RG}{2}},\;\; E = \frac{\mathcal{O}}{2G}.
\label{eq:width-connectivity}
\end{align}
This is illustrated in Figure \ref{fig:hg2}. As the group number $G$ increases, $C$ increases but $E$ decreases. 
The two curves intercept ($C=E$) when $G=\sqrt{C/R}$, at which each output channel connects to all input channels once. Mathematically, the resulting convolution matrix $\bm{W}$ is divided into $G \times G$ blocks, of which each has rank 1 (see Figure \ref{fig:low-rank}-Left).

\vspace{1mm}
\noindent \textbf{Micro-Factorized Depthwise Convolution:} 
As shown in Figure \ref{fig:low-rank}-Middle, we factorize a $k\times k$ depthwise convolution kernel into a $k\times1$ kernel and a $1\times k$ kernel. 
This shares the same mathematical format with Micro-Factorized pointwise convolution (Eq. \ref{eq:mat-fac}). The kernel matrix per channel $\bm{W}$ is of shape $k \times k$, which is decomposed into a $k \times 1$ vector $\bm{P}$ and a $1 \times k$ vector $\bm{Q^T}$. Here $\bm{\Phi}$ is a scalar with value 1. 
This low rank approximation reduces the computational complexity from $\mathcal{O}(k^2C)$ to $\mathcal{O}(kC)$.

\vspace{1mm}
\noindent \textbf{Combining Micro-Factorized Pointwise and Depthwise Convolutions:}
We combine Micro-Factorized pointwise and depthwise convolutions in two different ways: (a) regular combination, and (b) lite combination. The former simply concatenates the two convolutions. The lite combination uses a Micro-Factorized depthwise convolution to expand the number of channels by applying multiple spatial filters per channel. It then applies one group-adaptive convolution to fuse and squeeze the number of channels (shown in Figure \ref{fig:low-rank}-Right).
Compared to the regular counterpart, the lite combination is more effective at lower levels, as it saves computations from channel fusion (pointwise) to compensate for learning more spatial filters (depthwise).

\subsection{Dynamic Shift-Max}
Now we present dynamic Shift-Max, a new activation function to enhance non-linearity. It dynamically fuses an input feature map with its circular group shift, of which a group of channels are shifted. Dynamic Shift-Max also strengthens the connections between groups. This is complementary to Micro-Factorized pointwise convolution that focuses on connections within a group.

\begin{figure}[t]
	\begin{center}
		\includegraphics[width=0.80\linewidth]{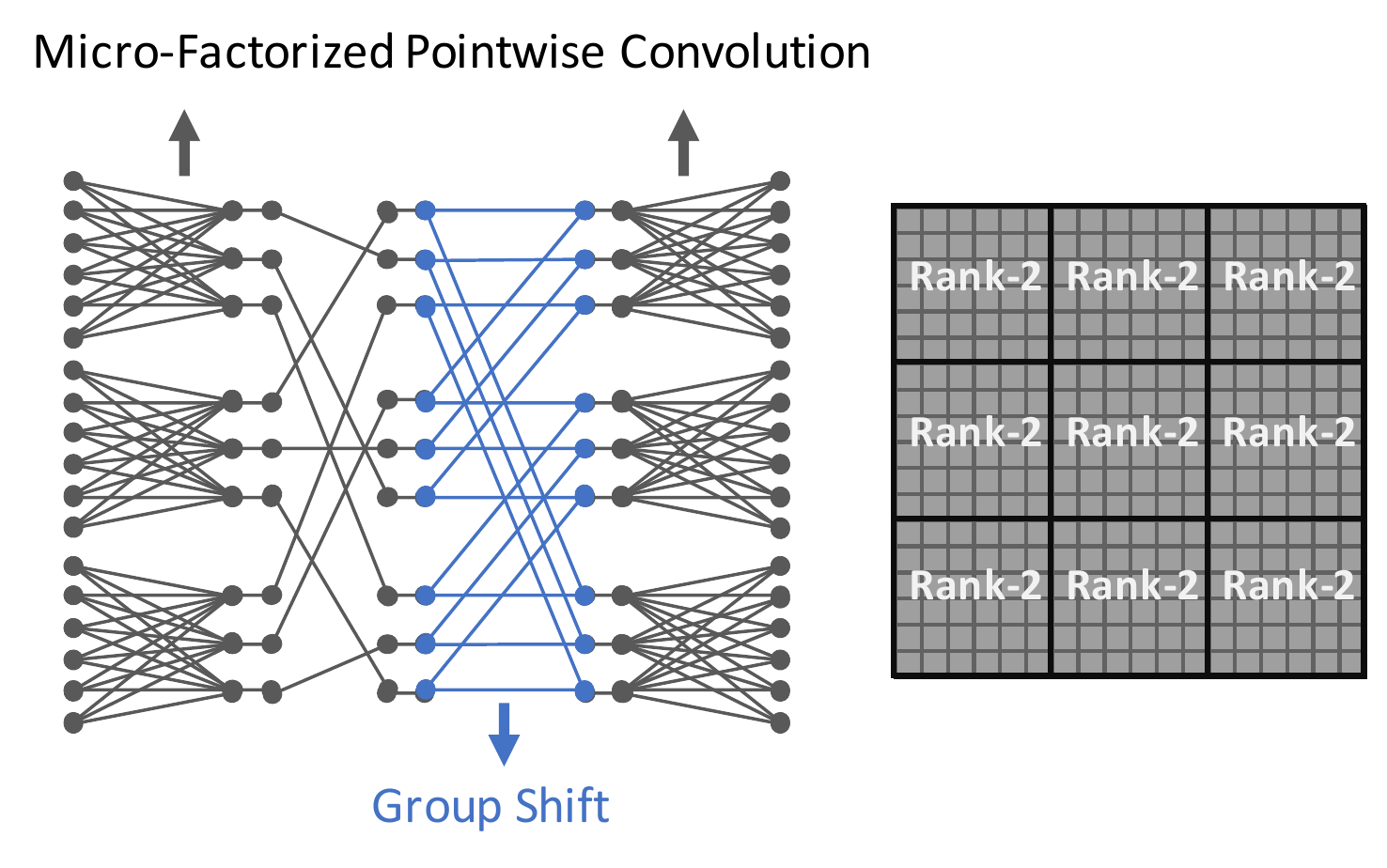}
	\end{center}
	\vspace{-3mm}
	\caption{\textbf{Group shift (a special case of dynamic Shift-Max)} improves Micro-Factorized pointwise convolution. The rank of each block in the resulting convolution matrix increases from 1 to 2.}
	\label{fig:dynamic-shift-max}
	\vspace{-3mm}
\end{figure}

\vspace{1mm}
\noindent \textbf{Definition:} Let $\bm{x}=\{x_i\}$ ($i=1,\dots,C$) denote an input vector (or tensor) with $C$ channels that are divided into $G$ groups. Each group has $\frac{C}{G}$ channels. Its $N$-channel circular shift can be represented as $x_N(i)=x_{(i+N) \bmod C}$. We extend channel shift to group shift by defining the group circular function as:
\begin{align}
x_{\frac{C}{G}}(i, j)=x_{(i+j\frac{C}{G}) \bmod C}, \; j=0,\dots,G-1,
\label{eq:group-circular}
\end{align}
where $x_{\frac{C}{G}}(i, j)$ corresponds to shifting the $i^{th}$ channel $x_i$ by $j$ groups. Dynamic Shift-Max combines multiple ($J$) group shifts as follows:
\begin{align}
y_i &= \max_{1 \leq k \leq K}\{ \sum_{j=0}^{J-1} a^k_{i, j}(\bm{x})x_{\frac{C}{G}}(i, j) \},
\label{eq:dynamic-group-shift-max}
\end{align}
where the parameter $a^k_{i,j}(\bm{x})$ adapts to the input $\bm{x}$ by a hyper function, which can be easily implemented by using two fully connected layers after average pooling, in a similar manner to Squeeze-and-Excitation \cite{squeezenet16}.

\vspace{1mm}
\noindent \textbf{Non-linearity:} Dynamic Shift-Max encodes two non-linearities: (a) it outputs the maximum of $K$ different fusions of $J$ groups, and (b) the parameter $a^k_{i,j}(\bm{x})$ is not static but a function of the input $\bm{x}$. These provide dynamic Shift-Max more representation power to compensate for the reduction of the number of layers. The recently proposed dynamic ReLU \cite{Chen2020DynamicReLU} is a special case of dynamic Shift-Max ($J=1$), where each channel is activated alone.

\vspace{1mm}
\noindent \textbf{Connectivity:}
Dynamic Shift-Max improves the connectivity between channel groups. It is complementary to Micro-Factorized pointwise convolution that focuses on connectivity within each group. Figure \ref{fig:dynamic-shift-max} shows that even a \textit{static group shift} ($y_i = a_{i,0}x_{\frac{C}{G}}(i,0)+a_{i,1}x_{\frac{C}{G}}(i, 1)$) can efficiently increase the rank for Micro-Factorized pointwise convolution. By inserting it between the two group-adaptive convolutions, the rank of each block in the resulting convolution matrix $\bm{W}$ ($G \times G$ block matrix) increases from 1 to 2. Note that the static group shift is a simple special case of dynamic Shift-Max with $K=1$, $J=2$, and static $a^k_{i,j}$.

\vspace{1mm}
\noindent \textbf{Computational Complexity:} Dynamic Shift-Max generates $CJK$ parameters $a^k_{i,j}(\bm{x})$ from input $\bm{x}$. The computational complexity includes three parts: (a) average pooling $\mathcal{O}(HWC)$, (b) generating parameters $a^k_{i,j}(\bm{x})$ in Eq. \ref{eq:dynamic-group-shift-max} $\mathcal{O}(C^2JK)$, and (c) applying dynamic Shift-Max per channel and per spatial location $\mathcal{O}(HWCJK)$. It is light-weight when $J$ and $K$ are small. Empirically, a good trade-off is achieved at $J=2$ and $K=2$. 

\subsection{Relation to Prior Work} 
MicroNet is related to the two popular efficient networks (MobileNet \cite{howard2017mobilenets, sandler2018mobilenetv2, Howard_2019_ICCV_mbnetv3} and ShuffleNet \cite{Zhang_2018_CVPR, ma_2018_ECCV}). It shares the reverse bottleneck structure with MobileNet and shares the using group convolution with ShuffleNet. In contrast, MicroNet differentiates from them in both convolutions and activation functions. Firstly, it factorizes pointwise convolution into group-adaptive convolutions, of which the group number adapts to the number of channels $G=\sqrt{C/R}$. Secondly, it factorizes depthwise convolution. Finally, a new activation (i.e dynamic Shift-Max) is proposed to improve both channel connectivity and non-linearity.  

\section{MicroNet Architecture} \label{section:micro-arch}
\begin{table*}[t!]
	\begin{center}
	    \footnotesize
	    \setlength{\tabcolsep}{1.9mm}{
		\begin{tabular}{@{\hskip 1mm}c|c@{\hskip 2.2mm}c@{\hskip 2.2mm}c@{\hskip 2.2mm}c@{\hskip 2.2mm}c|c@{\hskip 2.2mm}c@{\hskip 2.2mm}c@{\hskip 2.2mm}c@{\hskip 2.2mm}c|c@{\hskip 2.2mm}c@{\hskip 2.2mm}c@{\hskip 2.2mm}c@{\hskip 2.2mm}c|c@{\hskip 2.2mm}c@{\hskip 2.2mm}c@{\hskip 2.2mm}c@{\hskip 2.2mm}c@{\hskip 1mm}}
		    \specialrule{.1em}{.05em}{.05em} 
			& \multicolumn{5}{c|}{M0} & \multicolumn{5}{c|}{M1} & \multicolumn{5}{c|}{M2} & \multicolumn{5}{c}{M3}   \\
			\cline{2-21}
			Output & Block & $k$ & $C$ & $\frac{C}{R}$ & $G$ & Block & $k$ & $C$ & $\frac{C}{R}$ & $G$ & Block & $k$ & $C$ & $\frac{C}{R}$ & $G$ & Block & $k$ & $C$ & $\frac{C}{R}$ & $G$\\
		
			\specialrule{.1em}{.05em}{.05em} 
			112$\times$112 &  stem & 3 & 6 & 3 & (1, 3) &  stem & 3 & 8 & 4 & (1, 4) &  stem & 3 & 12 & 4 & (1, 4) &  stem & 3 & 16 & 4 & (1, 4)\\
			\hline
			56$\times$56 &  Micro-A & 3 & 24 & 8 & (2, --) &  Micro-A & 3 & 32 & 12 & (4,--) &  Micro-A & 3 & 48 & 16 & (4, --) &  Micro-A & 3 & 64 & 36 & (4, --)  \\
			\hline
			&  Micro-A & 3 & 32 & 16 & (4, --) &  Micro-A & 3 & 48 & 16 & (4,--) &  Micro-A & 3 & 64 & 24 & (4, -) &  Micro-B & 3 & 192 & 48 & (6, 8)  \\
			28$\times$28 &   &  &  &  &  & Micro-B & 3 & 144 & 24 & (4,6) & Micro-B & 3 & 144 & 24 & (4, 6) &  Micro-C & 3 & 192 & 48 & (6, 8)  \\
			&   &  &  &  &  &   &  &  &  &  &  & &  &  &  &  Micro-C & 3 & 192 & 48 & (6, 8)  \\
			\hline
			& Micro-B & 5 & 96 & 16 & (4,4) & Micro-C  & 5 & 192 & 32 & (4,8) &  Micro-C & 3 & 192 & 32 & (4,8) &  Micro-C & 5 & 256 & 64 & (8, 8)  \\
			& Micro-C & 5 & 192 & 32 & (4,8) & Micro-C  & 5 & 192 & 32 & (4,8) &  Micro-C & 5 & 192 & 32 & (4,8) &  Micro-C & 5 & 256 & 64 & (8, 8)  \\
			14$\times$14&  &  &  &  &  &  Micro-C & 5 & 384 & 64 & (8,8) & Micro-C & 5 & 288 & 48 & (6,8) &  Micro-C & 5 & 384 & 96 & (8, 12)  \\
			&  &  &  &  &  &   &  &  &  &  &  Micro-C & 5 & 480 & 80 & (8,10) &  Micro-C & 5 & 384 & 96 & (8, 12)  \\
			&  &  &  &  &  &   &  &  &  &  &  Micro-C & 5 & 480 & 80 & (8,10) &  Micro-C & 5 & 576 & 144 & (12, 12)  \\
			&  &  &  &  &  &   &  &  &  &  &  & &  &  &  &  Micro-C & 5 & 576 & 144 & (12, 12)  \\
			\hline
			& Micro-C & 5 & 384 & 64 & (8,8) &  Micro-C & 5 & 576 & 96 & (8,12) & Micro-C  & 5 & 720 & 120 & (10,12) &  Micro-C & 5 & 768 & 192 & (12, 16)  \\
			7$\times$7& Micro-C & 3 & 576 & 96 & (8,12) & Micro-C & 3 & 768 & 128 & (8,16) & Micro-C  & 3 & 720 & 120 & (10,12) &  Micro-C & 5 & 768 & 192 & (12, 16)  \\
			&  &  &  &  &  &  & &  &  &  &  Micro-C & 3 & 864 & 144 & (12,12) &  Micro-C & 5 & 768 & 192 & (12, 16)  \\
			&  &  &  &  &  &   &  &  &  &  &  & &  &  &  &  Micro-C & 3 & 1024 & 256 & (16, 16)  \\
			\hline
			1$\times$1&  \multicolumn{20}{c}{avg pool $\rightarrow$ 2fc $\rightarrow$ softmax}\\
			\hline
			 &  \multicolumn{5}{c|}{6M MAdds, 1.8M Param} & \multicolumn{5}{c|}{12M MAdds, 2.4M Param} & \multicolumn{5}{c|}{21M MAdds, 3.3M Param} & \multicolumn{5}{c}{44M MAdds, 4.5M Param} \\
			\specialrule{.1em}{.05em}{.05em} 
		\end{tabular}
		}
	\end{center}
	\vspace{-1mm}
	\caption{\textbf{MicroNet Architectures}. ``stem" refers to stem layer. ``Micro-A", ``Micro-B", and ``Micro-C" refers to three Micro-Blocks (see section \ref{section:micro-arch} and Figure \ref{fig:micro-block} for more details). $k$ is the kernel size, $C$ is the number of output channels, $R$ is the channel reduction ratio in Micro-Factorized pointwise convolution, $G$ is the group number. Note that for ``Micro-A" (see Figure \ref{fig:micro-block}a), $C$ is the number of output channels in Micro-Factorized depthwise convolution, $\frac{C}{R}$ is the number of output channels for the block.}
	\vspace{-1mm}
	\label{table:micro-arch}
\end{table*}

\begin{figure}[t]
	\begin{center}
		\includegraphics[width=1.0\linewidth]{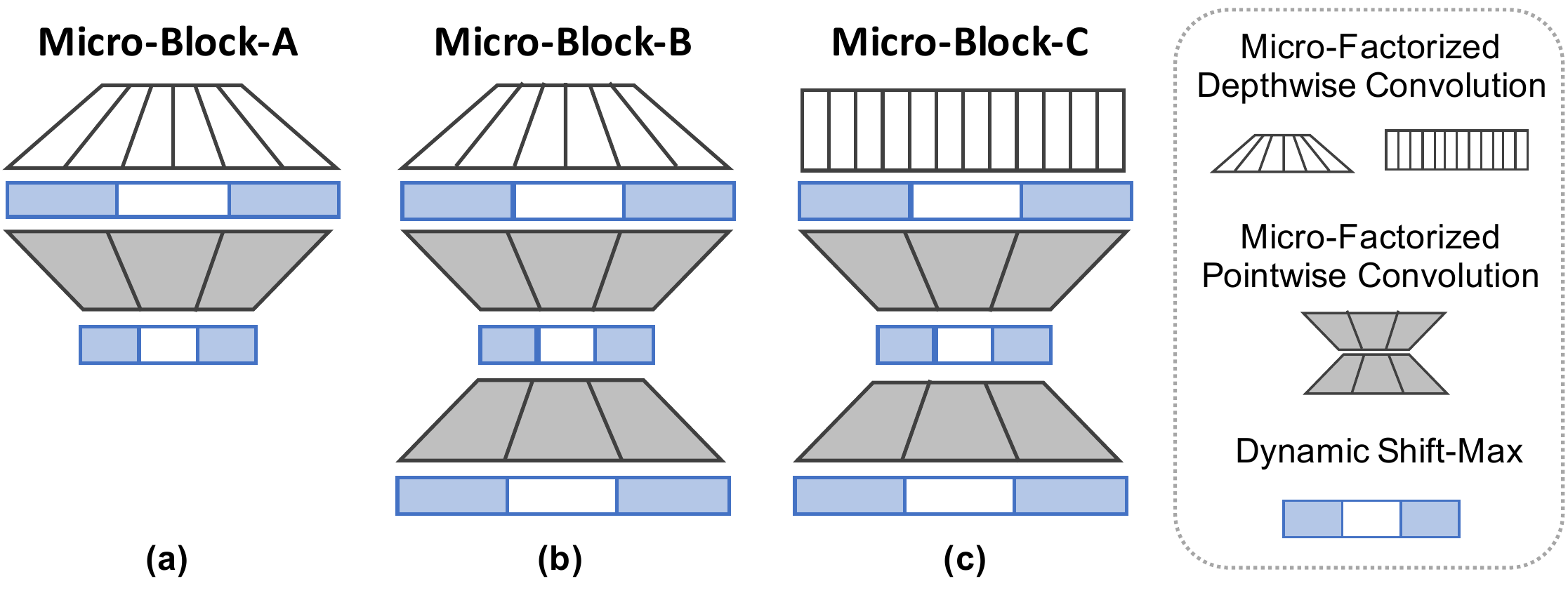}
	\end{center}
	\vspace{-2mm}
	\caption{\textbf{Diagram of three Micro-Blocks.} \textbf{(a) Micro-Block-A} that uses the lite combination of Micro-Factorized pointwise and depthwise convolutions (see Figure \ref{fig:low-rank}-Right). \textbf{(b) Micro-Block-B} that connects Micro-Block-A and Micro-Block-C. \textbf{(c) Micro-Block-C} that uses the regular combination of Micro-Factorized pointwise and depthwise convolutions. See Table \ref{table:micro-arch} for their usage.}
	\label{fig:micro-block}
	\vspace{-2mm}
\end{figure}
Now we describe the architectures of four MicroNet models, which have different FLOPs from 6M to 44M. They consist of three types of Micro-Blocks (see Figure \ref{fig:micro-block}), which combine Micro-Factorized pointwise and depthwise convolutions in different ways. All of them use dynamic Shift-Max as activation function. The details are listed as follows:

\vspace{1mm}
\noindent \textbf{Micro-Block-A:} As shown in Figure \ref{fig:micro-block}a, Micro-Block-A uses the lite combination of Micro-Factorized pointwise and depthwise convolutions (see Figure \ref{fig:low-rank}-Right). It is effective at lower levels that have higher resolution (e.g. $112 \times 112$ or $56 \times 56$). Note that the number of channels is expanded by Micro-Factorized depthwise convolution, and is squeezed by using a group-adaptive convolution. 

\vspace{1mm}
\noindent \textbf{Micro-Block-B:} 
Micro-Block-B is used to connect Micro-Block-A and Micro-Block-C. Different to Micro-Block-A, it uses a full Micro-Factorized pointwise convolution that includes two group-adaptive convolutions (shown in Figure \ref{fig:micro-block}b). The former squeezes the number of channels, but the latter expands the number of channels. Each MicroNet only has one Micro-Block-B (see Table \ref{table:micro-arch}). 

\vspace{1mm}
\noindent \textbf{Micro-Block-C:} Micro-Block-C (shown in Figure \ref{fig:micro-block}c) uses the regular combination that concatenates Micro-Factorized depthwise and pointwise convolutions. It is used at higher levels (see Table \ref{table:micro-arch}) as it spends more computations on channel fusion (pointwise) than the lite combination. The skip connection is used when the dimension matches.

Each micro-block has four hyper-parameters: kernel size $k$, number of output channels $C$, reduction ratio in the bottleneck of Micro-Factorized pointwise convolution $R$, and group number pair $(G_1, G_2)$ for the two group-adaptive convolutions. Note that we relax Eq. \ref{eq:g-cr} as $G_1G_2=C/R$ and find the close integer solution.

\vspace{1mm}
\noindent \textbf{Stem Layer:} we redesign the stem layer to meet the low FLOP constraint. It includes a $3 \times 1$ convolution and a $1 \times 3$ group convolution, and is followed by a ReLU. The second convolution expands the number of channels by $R$ times. This substantially saves the computational cost. For instance, the stem layer in MicroNet-M3 (see Table \ref{table:micro-arch}) only need 1.5M MAdds.

\vspace{1mm}
\noindent \textbf{Four MicroNet Models (M0--M3):} We design four models (M0, M1, M2, M3) with different computational costs (6M, 12M, 21M, 44M MAdds). Table \ref{table:micro-arch} shows their full specification.
These networks follow the same pattern from low to high levels: stem layer $\rightarrow$ Micro-Block-A $\rightarrow$ Micro-Block-B $\rightarrow$ Micro-Block-C. Note that all models are manually designed without network architecture search (NAS).

\section{Experiments: ImageNet Classification}
Below we evaluate four MicroNet models (M0--M3) along with comprehensive ablations on ImageNet \cite{deng2009imagenet} classification. ImageNet has 1000 classes, including 1,281,167 images for training and 50,000 images for validation. 

\subsection{Implementation Details} \label{sec:impl-details}

\vspace{1mm}
\noindent \textbf{Training Strategy:} Each model is trained in two ways: (a) stand alone, and (b) mutual learning. The former is straightforward that the model learns by itself. The latter co-learns a full rank partner along each MicroNet  where the full rank partner shares the same network width/height, but replaces Micro-Factorized pointwise and depthwise convolutions with the original pointwise and depthwise ($k \times k$) convolutions. KL divergence is used to encourage MicroNet to learn from its corresponding full rank partner.  

 \vspace{1mm}
\noindent \textbf{Training Setup:} All models are trained using SGD optimizer with 0.9 momentum. The image resolution is 224$\times$224. We use a mini-batch size of 512, and a learning rate of 0.02. Each model is trained for 600 epochs with cosine learning rate decay. The weight decay is 3e-5 and dropout rate is $0.05$ for smaller MicroNets (M0 and M1). For larger models (M2 and M3), the weight decay is 4e-5 and dropout rate is $0.1$. Label-smoothing ($0.1$) and Mixup \cite{zhang2018mixup} ($0.2$) is used for MicroNet-M3 to avoid overfitting.

\subsection{Main Results}
Table \ref{table:imagenet-cls-result} compares MicroNets with the state-of-the-art on ImgageNet classification at four different computational costs. MicroNets outperform all prior works across all four FLOP constraints by a clear margin. For instance, without mutual learning, MicroNets outperform MobileNetV3 by 9.6\%, 9.6\%, 6.1\% and 4.4\% at 6M, 12M, 21M and 44M FLOPs, respectively. When trained with mutual learning, all four MicroNets gain about 1.5\% top-1 accuracy consistently. Our method achieves 53.0\% top-1 accuracy at 6M FLOPs, outperforming MobileNetV3 with doubled complexity (12M FLOPs) by 3.2\%.
When comparing with the recent improvement over MobileNet and ShuffleNet, like GhostNet \cite{Han_2020_CVPR_ghostnet}, WeightNet \cite{Ma_2020_eccv_WeightNetRT} and ButterflyTransform \cite{vahid_2020_CVPR}, our method gains more than 5\% top-1 accuracy with similar FLOPs.
This demonstrates that MicroNet is effective to handle the extremely low FLOPs.

\begin{table}[t!]
	\begin{center}
	    \footnotesize
		\begin{tabular}{l@{\hskip 2.5mm}|c@{\hskip 2.5mm}r@{\hskip 2.5mm}|c@{\hskip 2.5mm}c@{\hskip 2mm}}
		    \specialrule{.1em}{.05em}{.05em} 
			Model & \#Param & MAdds &Top-1 &	Top-5   \\
		
			\specialrule{.1em}{.05em}{.05em} 
			MobileNetV3 $0.2 \times^{\dag}$ & 1.2M & 6M & 41.1 & 65.2	  \\
			MobileNetV3 $0.25 \times^{\dag}$ & 1.2M & 6M & 41.8 & 65.8	  \\
			\textbf{MicroNet-M0}  & 1.8M & 6M &  51.4 & 74.5  \\
			\textbf{MicroNet-M0 (ML)}  & 1.8M & 6M &  \textbf{53.0} &  \textbf{76.0} \\
			
			\hline
			ShuffleNetV1 $0.25 \times$ \cite{Zhang_2018_CVPR} & -- & 13M & 47.3 & -- \\
		    MobileNetV3 $0.35 \times$ \cite{Howard_2019_ICCV_mbnetv3} & 1.4M & 12M  & 49.8 & -- 		 \\
		    MobileNetV3+BFT $0.5 \times$ \cite{vahid_2020_CVPR} & -- & 15M & 55.2 & -- \\
		    \textbf{MicroNet-M1}  & 2.4M  & 12M & 59.4 & 80.9 		 \\
			\textbf{MicroNet-M1(ML)}  & 2.4M & 12M & \textbf{61.1} &  \textbf{82.4}		 \\
			\hline
			ShuffleNetV2+BFT \cite{vahid_2020_CVPR} & -- & 21M & 57.8 & -- \\
			MobileNetV3 $0.5 \times$ \cite{Howard_2019_ICCV_mbnetv3} & 1.6M & 21M & 58.0 & 	--	 \\
			\textbf{MicroNet-M2}  & 3.3M & 21M & 64.1 &  84.5		 \\
			\textbf{MicroNet-M2 (ML)}  & 3.3M & 21M & \textbf{65.7} & \textbf{86.0} 		 \\
			\hline
		    ShuffleNetV2 $0.5 \times$ \cite{ma_2018_ECCV} & 1.4M & 41M & 60.3 & --  		 \\
			MobileNetV3 $0.75 \times$ \cite{Howard_2019_ICCV_mbnetv3} & 2.0M & 44M & 65.4 & --  		 \\
			GhostNet $0.5 \times$ \cite{Han_2020_CVPR_ghostnet} & 2.6M & 42M & 66.2 & 86.6 \\ 
			WeightNet $8 \times$ \cite{Ma_2020_eccv_WeightNetRT} & 2.7M & 42M & 66.0 & -- \\
			\textbf{MicroNet-M3} &4.5M & 44M & 69.8 & 	88.3	 \\
			\textbf{MicroNet-M3 (ML)} &4.5M & 44M & \textbf{71.2} & \textbf{89.6}		 \\
			\specialrule{.1em}{.05em}{.05em} 
		\end{tabular}
	\end{center}
	\caption{\textbf{ImageNet \cite{deng2009imagenet} classification results}. ``ML" stands for mutual learning with a full rank partner (see Section \ref{sec:impl-details}). $^{\dag}$ indicates our implementation. MobileNetV3 $0.2 \times$ only reduces network width, while MobileNetV3 $0.25 \times$ reduces both width and depth (removing two blocks). ``--": not available in the original paper.}
	\label{table:imagenet-cls-result}
	\vspace{-2mm}
\end{table}

\subsection{Ablation Studies}
We run a number of ablations to analyze MicroNet. MicroNet-M1 (12M FLOPs) is used for all ablations, and each model is trained for 300 epochs. The default hyper parameters for dynamic Shift-Max are set as $J=2$, $K=2$.

\begin{table}[t!]
	\begin{center}
	    \footnotesize
	    \setlength{\tabcolsep}{1.7mm}{
		\begin{tabular}{c@{\hskip 2mm}|@{\hskip 1mm}c@{\hskip 2mm}c@{\hskip 2mm}c@{\hskip 1mm}|@{\hskip 1mm}c@{\hskip 2mm}c@{\hskip 1mm}|@{\hskip 1mm}c@{\hskip 2mm}c@{\hskip 1mm}|@{\hskip 1mm}c}
		    \specialrule{.1em}{.05em}{.05em}
		    & \multicolumn{3}{@{\hskip 1mm}c|@{\hskip 1mm}}{Micro-Fac Conv} & \multicolumn{2}{@{\hskip 1mm}c|@{\hskip 1mm}}{Shift-Max} & & & \\
		    & DW & PW & Lite & static & dynamic & Param & MAdds & Top-1 \\
		   \specialrule{.1em}{.05em}{.05em}
		    Mobile & & & & & &1.3M &10.6M &44.9 \\
		    \hline
		    &\checkmark  &  & & & & 1.7M & 10.6M & 46.4\\
		    &\checkmark &\checkmark & & & &1.7M & 10.6M &50.0 \\
            Micro&\checkmark & \checkmark & \checkmark & & & 1.8M & 10.5M & 51.7 \\
            & \checkmark & \checkmark & \checkmark & \checkmark & & 1.9M & 11.8M & 54.4 \\
            & \checkmark & \checkmark & \checkmark &  & \checkmark & 2.4M & 12.4M & \textbf{58.5} \\
	       \specialrule{.1em}{.05em}{.05em}
		\end{tabular}
		}
	\end{center}
	\caption{\textbf{The path from MobileNet to MicroNet} evaluated on ImageNet classification. Here, we modify MobileNet-V2 such that it has similar FLOPs (about 10.6M) to three Micro-Factorized convolution options: depthwise (DW), pointwise (PW), and lite combination at low levels (Lite). We also compare dynamic Shift-Max with its static counterpart (static $a^k_{i,j}$ in Eq. \ref{eq:dynamic-group-shift-max}).
	}
	\vspace{-2mm}
	\label{table:mobile-to-micro}
\end{table}

\vspace{1mm}
\noindent \textbf{From MobileNet to MicroNet:} Table \ref{table:mobile-to-micro} shows the path from MobileNet to our MicroNet. Both share the reverse bottleneck structure. Here, we modify MobileNetV2 \cite{sandler2018mobilenetv2} (without SE \cite{Hu_2018_CVPR}) such that it has similar complexity (10.5M MAdds) with three Micro-Factorized convolution variations (row 2--4). Micro-Factorized pointwise and depthwise convolutions and their lite combination at low levels boost the top-1 accuracy from 44.9\% to 51.7\% step by step.
Furthermore, using static and dynamic Shift-Max gains another 2.7\% and 6.8\% top-1 accuracy respectively, with a small amount of additional cost. This demonstrates that the proposed \textit{Micro-Factorized Convolution} and \textit{Dynamic Shift-Max} are effective and complementary to handle extremely low computational cost. 

\begin{table*}[t]
\parbox{.33\linewidth}{
    \begin{center}
	    \footnotesize
		\begin{tabular}{c@{\hskip 2.5mm}|c@{\hskip 2.5mm}c@{\hskip 2.5mm}c@{\hskip 2.5mm}}
		    \specialrule{.1em}{.05em}{.05em} 
			$G$ & Param & MAdds  & Top-1  \\[0.5em]
			\specialrule{.1em}{.05em}{.05em} 
			 1 & 1.3M & 10.6M  & 48.8  \\
			 2 & 1.5M & 10.5M  & 50.2 \\
			 4 & 1.7M & 10.6M & 50.7 \\
			 8 & 1.7M & 10.6M & 50.8 \\
			\hline
            $G=\sqrt{C/R}$ & 1.8M & 10.5M  & \textbf{51.7} \\
			\specialrule{.1em}{.05em}{.05em} 
			\multicolumn{4}{c}{} \\[-0.5em]
			\multicolumn{4}{c}{(a) \textbf{Fixed group number $G$.}} \\
		\end{tabular}
	\end{center}
	\label{table:group-number-1}
	\vspace{-0.5em}
}
\hfill
\parbox{.33\linewidth}{
    \begin{center}
	    \footnotesize
		\begin{tabular}{c@{\hskip 2.5mm}|r@{\hskip 2.5mm}r@{\hskip 2.5mm}r@{\hskip 2.5mm}}
		    \specialrule{.1em}{.05em}{.05em} 
			$\lambda=\frac{G}{\sqrt{C/R}}$ & Param & MAdds  & Top-1  \\
			\specialrule{.1em}{.05em}{.05em} 
			$\;\;\;$ $0.25$ & 1.5M & 10.5M   & 50.2    \\
			$\;\;\;\;$ $0.5$ & 1.7M & 10.6M  & 51.6 \\
            \xmark$\:$ $1.0$ & 1.8M & 10.5M  & \textbf{51.7} \\
            $\;\;\;\;$ $2.0$ & 2.1M & 10.5M  & 50.6 \\
            $\;\;\;\;$ $4.0$ & 2.2M & 10.7M & 47.6 \\
			\specialrule{.1em}{.05em}{.05em} 
			\multicolumn{4}{c}{}\\[-0.5em]
			\multicolumn{4}{c}{(b) \textbf{Adaptive group number $G$.}} \\
		\end{tabular}
	\end{center}
	\label{table:group-number-2}
	\vspace{-0.5em}
}
\hfill
\parbox{.33\linewidth}{
    \begin{center}
	    \footnotesize
		\begin{tabular}{r@{\hskip 2.5mm}c|c@{\hskip 2.5mm}c@{\hskip 2.5mm}c}
		    \specialrule{.1em}{.05em}{.05em}
		    \multicolumn{2}{c|}{Levels} & & & \\
		    low & high & Param & MAdds  & Top-1  \\
		   \specialrule{.1em}{.05em}{.05em}
		     & &1.7M &10.6M &  50.0 \\
		     \xmark$\:$ \checkmark & & 1.8M &10.5M & \textbf{51.7} \\
		     \checkmark & \checkmark &2.0M & 10.6M & 51.2 \\
	       \specialrule{.1em}{.05em}{.05em}
	       \multicolumn{5}{c}{}\\
	       \multicolumn{5}{c}{}\\[-0.5em]
	       \multicolumn{5}{c}{}\\[-0.5em]
		\multicolumn{5}{c}{(c) \textbf{Lite combination} at different levels} \\
		\end{tabular}
	\end{center}
    \label{table:half-and-half}
	\vspace{-0.5em}
}
\caption{\textbf{Ablations of Micro-Factorized convolution} on ImageNet classification. \xmark $\:$ indicates the default choice for the rest of the paper.}
\label{table:mfc-pointwise}
\vspace{-3mm}
\end{table*}

\vspace{1mm}
\noindent \textbf{Group number $G$:}
Micro-Factorized pointwise convolution includes two group-adaptive convolutions, where their group numbers are selected by relaxing $G=\sqrt{C/R}$ to close integers. Table \ref{table:mfc-pointwise}a compares it with networks that have similar structure and FLOPs (about 10.5M MAdds), but use a fixed group number. Group-adaptive convolution achieves higher accuracy, demonstrating a good balance between the number of channels and input/output connectivity.

Table \ref{table:mfc-pointwise}b compares different options of adaptive group number, which are controlled by a multiplier $\lambda$ such that $G=\lambda \sqrt{C/R}$. The bigger value of $\lambda$ corresponds to more channels but less input/output connectivity (see Figure \ref{fig:hg2}). The good balance is achieved when $\lambda$ is between 0.5 and 1. The top-1 accuracy drops when $\lambda$ either increases (more channels but less connectivity) or decreases (fewer channels but more connectivity). Thus, we use $\lambda=1$ for the rest of the paper. Please note that all models in Table \ref{table:mfc-pointwise}b have similar computational costs (about 10.5M MAdds).

\vspace{1mm}
\noindent \textbf{Lite combination at different levels:} Table \ref{table:mfc-pointwise}c compares using the lite combination of Micro-Factorized pointwise and depthwise convolutions (see Figure \ref{fig:low-rank}-Right) at different levels. Using it at low levels alone achieves the highest accuracy. This validates that the lite combination is more effective at lower levels. Compared to the regular combination, it saves computations from channel fusion (pointwise) to compensate for learning more spatial filters (depthwise).

\vspace{1mm}
\noindent \textbf{Comparing with other activation functions:} We compare dynamic Shift-Max with three existing activation functions ReLU \cite{NairH10Relu}, SE+ReLU \cite{Hu_2018_CVPR}, and dynamic ReLU \cite{Chen2020DynamicReLU}. The results are shown in Tabel \ref{table:ablation-dy-activation}. Our dynamic Shift-Max outperforms the other three by a clear margin (2.5\%), demonstrating its superiority. Note that dynamic ReLU is a special case of dynamic Shift-Max with $J=1$ (see Eq. \ref{eq:dynamic-group-shift-max}). 

\begin{table}[t!]
	\begin{center}
	    \footnotesize
	    \setlength{\tabcolsep}{3.1mm}{
		\begin{tabular}{l|rr|ll}
		    \specialrule{.1em}{.05em}{.05em} 
			Activation & Param & MAdds &Top-1 &	Top-5   \\
		
			\specialrule{.1em}{.05em}{.05em} 
			ReLU\cite{NairH10Relu} & 1.8M & 10.5M & 51.7  & 74.3 \\
			SE\cite{Hu_2018_CVPR}+ReLU & 2.1M & 10.9M & 54.4 & 76.8 \\
			Dynamic ReLU \cite{Chen2020DynamicReLU} & 2.4M & 11.8M & 56.0 & 78.0 \\
			\hline
			Dynamic Shift-Max & 2.4M & 12.4M & \textbf{58.5} & \textbf{80.1} \\
			\specialrule{.1em}{.05em}{.05em} 
		\end{tabular}
		}
	\end{center}
	\caption{\textbf{Dynamic Shift-Max vs. other activation functions} on ImageNet classification. MicroNet-M1 is used.}
	\label{table:ablation-dy-activation}
\end{table}

\vspace{1mm}
\noindent \textbf{Dynamic Shift-Max at different layers:}
Table \ref{table:ablation-diff-layer} shows the top-1 accuracy for using dynamic Shift-Max at three different layers in a micro-block (see Figure \ref{fig:micro-block}). Using it at more layers results in consistent improvement. The best accuracy is achieved when using it for all three layers. If only one layer is allowed to use dynamic Shift-Max, the recommendation is to use it after the depthwise convolution. 

\begin{table}[t!]
	    \footnotesize
	    \setlength{\tabcolsep}{2.7mm}{
        \begin{tabular}{c|c@{\hskip 2.5mm}c@{\hskip 2.5mm}c|c@{\hskip 2.5mm}c|c@{\hskip 2.5mm}c}
        \specialrule{.1em}{.05em}{.05em}
         & $A_1$ & $A_2$ & $A_3$ & Param & MAdds & Top-1 & Top-5 \\
        \specialrule{.1em}{.05em}{.05em}
        ReLU & -- & -- & -- &1.8M & 10.5M&51.7 & 74.3 \\
        \hline
         & \checkmark & -- & -- & 2.1M& 11.3M&55.9 & 77.9 \\
         & -- & \checkmark & -- &2.0M  & 10.6M&53.3 & 76.0 \\
        Dynamic & -- & -- & \checkmark& 2.1M &11.2M & 54.8 & 77.2 \\
        Shift-Max & \checkmark & \checkmark & -- & 2.2M&11.5M &56.6& 78.3\\
         & \checkmark & -- & \checkmark & 2.3M &12.2M &57.9 & 79.6\\
         & -- & \checkmark & \checkmark &2.2M &11.4M &55.5 & 77.8 \\        
         & \checkmark & \checkmark & \checkmark &2.4M &12.4M & \textbf{58.5} & \textbf{80.1} \\

        \specialrule{.1em}{.05em}{.05em}
         \multicolumn{6}{c}{}\\
        \end{tabular}
        }
        \caption{\textbf{Dynamic Shift-Max at different layers} evaluated on ImageNet. MicroNet-M1 is used. $A_1,A_2,A_3$ indicate three activation layers sequentially in Micro-Block-B and Micro-Block-C (see Figure \ref{fig:micro-block}). Micro-Block-A only includes $A_1$ and $A_2$.}
        \label{table:ablation-diff-layer}
	\vspace{-3mm}
\end{table}

\vspace{1mm}
\noindent \textbf{Different hyper parameters in dynamic Shift-Max:} Table \ref{table:ablation-J-K} shows the results of using different combinations of $K$ and $J$ (in Eq. \ref{eq:dynamic-group-shift-max}). We add ReLU when $K=1$ as only one element is left in the max operator. The baseline at the first row ($J=1$, $K=1$) is equivalent to SE+ReLU \cite{Hu_2018_CVPR}.
When fixing $J=2$ (fusing two groups), the winner of two fusions ($K=2$) is better than a single fusion ($K=1$). Adding the third fusion is not helpful, since it is covered mostly by the other two fusions but involves more parameters. When fixing $K=2$ (maximum of two fusions), involving more groups $J$ is consistently better but introduces more FLOPs. A good tradeoff is achieved at $J=2$ and $K=2$, where 4.1\% gain is achieved with additional 1.5M MAdds.

\begin{table}[t!]
	\begin{center}
	    \footnotesize
	    \setlength{\tabcolsep}{3.7mm}{
		\begin{tabular}{rc|cc|cc}
		    \specialrule{.1em}{.05em}{.05em} 
			$J$ & $K$ & Param & MAdds &Top-1 &	Top-5   \\
		
			\specialrule{.1em}{.05em}{.05em} 
			1 & 1 & 2.1M & 10.9M & 54.4  & 76.8 \\
			 \hline
			2 & 1 & 2.2M & 11.8M & 55.9 & 78.2 \\
			\xmark$\:$ 2 & 2 & 2.4M & 12.4M & 58.5 & 80.1 \\
			2 & 3 & 2.6M & 13.8M& 58.1 & 79.7 \\
			\hline
			1 & 2 & 2.2M & 11.2M & 55.5 & 77.6 \\
			\xmark$\:$ 2 & 2 & 2.4M & 12.4M & 58.5 & 80.1 \\
			3 & 2 & 2.6M & 14.2M & 59.0 & \textbf{80.3} \\
			\hline
			3 & 3 & 2.8M & 15.3M & \textbf{59.1} & \textbf{80.3} \\
			\specialrule{.1em}{.05em}{.05em} 
		\end{tabular}
		}
	\end{center}
	\caption{\textbf{Ablations of two hyper parameters in dynamic Shift-Max} ($J$, $K$ in Eq. \ref{eq:dynamic-group-shift-max}) on ImageNet classification. \xmark $\:$ indicates the default choice for the rest of the paper.}
	\label{table:ablation-J-K}
	\vspace{-3mm}
\end{table}

\section{MicroNet for Pixel-Level Classification}
MicroNet is not only effective for the image-level classification, but also works well for pixel-level tasks. In this section, we will show its application in human pose estimation and semantic segmentation.

\subsection{Human Pose Estimation}
\begin{table*}[t]
	\begin{center}
		\footnotesize
		\setlength{\tabcolsep}{2.8mm}{
		\begin{tabular}{ll| r r| l c c c c}
			\specialrule{.1em}{.05em}{.05em} 
			Backbone & Head & Param & MAdds & AP &	AP$^{0.5}$ & AP$^{0.75}$ & AP$^M$ & AP$^L$\\
			\specialrule{.1em}{.05em}{.05em}
			MobileNetV2 $\times0.5$ +DY-Conv \cite{Chen2019DynamicCA} & Mobile-Blocks + DY-Conv  & 10.0M & 807.4M & {62.8}         & {86.1} & 	{70.4} &	{59.9}&		{68.6}\\
			MobileNetV2 $\times0.5$ + DY-ReLU \cite{Chen2020DynamicReLU} & Mobile-Blocks + DY-ReLU& 4.6M & 820.3M & 63.3 & 	86.3 &	71.4&		60.3&	69.2\\
			
			\hline
			MobileNetV3 Small + DY-Conv \cite{Chen2019DynamicCA} &  Mobile-Blocks + DY-Conv    & 7.7M & 716.2M & 60.0 & 85.0 & 67.8 &	57.6&		65.4\\
			MobileNetV3 Small + DY-ReLU \cite{Chen2020DynamicReLU} & Mobile-Blocks + DY-ReLU& 4.8M & 747.9M & 60.7 & 85.7	 &	68.1 &	58.1	&	66.3\\	
			\hline
			\textbf{MicroNet-M3} & Micro-Blocks & 4.0M & \textbf{263.2M} & 62.8 & 86.2 & 70.6& 60.0 & 68.4 \\
			\textbf{MicroNet-M2} & Micro-Blocks & 2.2M & \textbf{163.2M} & 58.7 & 84.0 & 65.5& 56.0 & 64.2 \\
			\textbf{MicroNet-M1} & Micro-Blocks & 1.8M & \textbf{116.8M} & 54.9 & 82.0 & 60.3  & 53.2 & 59.6\\
			\textbf{MicroNet-M0} & Micro-Blocks & 1.0M & \textbf{77.7M} & 50.3 & 79.6 & 53.9 & 48.3 & 54.8 \\
			\specialrule{.1em}{.05em}{.05em} 
		\end{tabular}
		}
	\end{center}
	\caption{\textbf{COCO keypoint detection results}. MicroNets are compared to two dynamic networks (dynamic convolution \cite{Chen2019DynamicCA} and dynamic ReLU \cite{Chen2020DynamicReLU}), which are built upon MobileNet V2 \cite{sandler2018mobilenetv2} and V3 \cite{Howard_2019_ICCV_mbnetv3} and share the same structure in both backbone and head.
	}
	\vspace{-2mm}
	\label{table:coco-kp}
\end{table*}

We use COCO 2017 dataset \cite{lin2014microsoft} to evaluate MicroNet on single-person keypoint detection. Our models are trained on \texttt{train2017}, including $57K$ images and $150K$ person instances labeled with 17 keypoints. We evaluate our method on \texttt{val2017} containing 5000 images and use the mean average precision (AP) over 10 object key point similarity (OKS) thresholds as the metric.

\vspace{1mm}
\noindent \textbf{Implementation Details:} Similar to image classification, we have four MicroNet models (M0--M3) for keypoint detection with different FLOPs. The models are modified to fit the keypoint detection task, by increasing resolution ($\times 2$) for a set of selected blocks (e.g. all blocks with stride of 32). The selection varies for different MicroNet models
 (see appendix \ref{apx:arch-kp} for more details). Each model has a head that includes three micro blocks (one with stride of 8 and two with stride of 4) and a pointwise convolution to generate heatmaps for 17 keypoints. We use bilinear upsampling to increase resolution in the head, and use spatial attention \cite{Chen2020DynamicReLU} per layer.

\vspace{1mm}
\noindent \textbf{Training Setup:} The training setup in \cite{sun2019deep} is used. The human detection boxes are cropped and resized to $256\times192$. The data augmentation includes random rotation ($[-\ang{45}, \ang{45}]$), random scale ($[0.65, 1.35]$), flipping, and half body data augmentation. All models are trained from scratch for 250 epochs, using Adam optimizer \cite{kingma:adam}. The initial learning rate is set as 1e-3 and is dropped to 1e-4 and 1e-5 at the $210^{th}$ and $240^{th}$ epoch, respectively.

\noindent \textbf{Testing:} The two-stage top-down paradigm \cite{xiao2018simplebaseline, sun2019deep} is used for testing: detecting person instances and then predicting keypoints. We use the same person detectors provided by \cite{xiao2018simplebaseline}.
The heatmaps of the original and flipped images are combined, on which the keypoints are predicted by adjusting the highest heat value location with a quarter offset towards the second highest response.

\noindent \textbf{Main Results:}
Table \ref{table:coco-kp} compares MicroNets with prior works \cite{Chen2020DynamicReLU, Chen2019DynamicCA} on efficient pose estimation, of which the computational cost is less than 850 MFLOPs. Both works use MobileNet’s inverted residual bottleneck
blocks in both backbone and head, and show clear improvement by adapting parameters in convolutions \cite{Chen2019DynamicCA} and activation functions \cite{Chen2020DynamicReLU} to the input. 
Our MicroNet-M3 only consumes 33\% of FLOPs in these works but achieves similar performance, demonstrating our method is also effective for keypoint detection. Furthermore, MicroNet-M2, M1, M0 provide good baselines for keypoint detection with even lower computational complexity ranging from 77M to 163M FLOPs.

\subsection{Semantic Segmentation}
\begin{table}[t!]
	\begin{center}
	    \footnotesize
	    \setlength{\tabcolsep}{2.7mm}{
		\begin{tabular}{l|l|cc|c}
		    \specialrule{.1em}{.05em}{.05em} 
			Backbone& Head & Param & MAdds & mIOU   \\
		
			\specialrule{.1em}{.05em}{.05em} 
			MBNetV2 0.5 & LR-ASPP & 0.28M & 4.00B & 68.6 \\
			MBNetV3-Small & LR-ASPP & 0.47M & 2.90B & 68.4 \\
			\hline
			\textbf{MicroNet-M3}  & MR-ASPP  & 1.87M &  \textbf{2.52B} & \textbf{69.1}  \\
			\textbf{MicroNet-M2}  & MR-ASPP & 1.85M & \textbf{1.75B}  &  66.1 \\
			\textbf{MicroNet-M1}  & MR-ASPP & 0.99M & \textbf{1.20B}  & 63.5  \\
			\textbf{MicroNet-M0}  & MR-ASPP & 0.43M &  \textbf{0.81B} &  56.9 \\
			\specialrule{.1em}{.05em}{.05em} 
		\end{tabular}
		}
	\end{center}
	\caption{\textbf{Semantic segmentation results} on Cityscapes validation set. The results in the first two rows are from \cite{Howard_2019_ICCV_mbnetv3}.}
	\vspace{-2mm}
	\label{table:city-seg-result}
\end{table}

We conduct experiments on Cityscape dataset \cite{Cordts2016Cityscapes} with fine annotations to evaluate MicroNet on semantic segmentation. Our models are trained on \texttt{train\_fine} set, including 2,975 images. We evaluate our method on \texttt{val} set containing 500 images, and use mIOU as the metric.

\vspace{1mm}
\noindent \textbf{Implementation Details:}
We modify four MicroNet models (M0--M3) as the backbone by increasing the resolution for all blocks with stride of 32 to stride of 16, similar to MobileNetV3 \cite{Howard_2019_ICCV_mbnetv3}. Our models have very low computational cost from 2.5B to 0.8B FLOPs on image resolution 1024$\times$2048. We follow the Lite Reduced design of Atrous Spatial Pyramid Pooling (LR-ASPP) \cite{Howard_2019_ICCV_mbnetv3} in the segmentation head that bilinearly upsamples the feature map by 2, applies spatial attention, and merges with the feature map from the backbone at stride of 8. We make LR-ASPP lighter by using Micro-Factorized convolutions to replace $1 \times 1$ convolution, denoted as Micro-Reduce ASPP (MR-ASPP). 

\noindent \textbf{Training Setup:} All models are randomly initialized and trained for $240$ epochs. The initial learning rate is set as $0.2$, and is decayed to 1e-4 with a cosine function. The weight decay is set as 4e-5. The data augmentation in \cite{Chen2017RethinkingAC} is used.

\noindent \textbf{Main Results:}
Table \ref{table:city-seg-result} reports the mIOU for all four MicroNets. Compared to MobileNetV3 (68.4 mIOU with 2.90B MAdds), our MicroNet-M3 is more accurate (69.1 mIOU) with lower computational cost (2.52B MAdds). This demonstrates the superiority of our methods in semantic segmentation. In addition, our MicroNet-M2, M1, M0 provide good baselines for semantic segmentation with even lower FLOPs from 1.75B to 0.81B MAdds.

\section{Conclusion}
In this paper, we presented MicroNet to handle extremely low computational cost. It builds on two proposed operators: Micro-Factorized convolution and Dynamic Shift-Max. The former balances between the number of channels and input/output connectivity via low rank approximations on both pointwise and depthwise convolutions. The latter fuses consecutive channel groups dynamically, enhancing both node connectivity and non-linearity to compensate for the depth reduction. A family of MicroNets achieve solid improvement for three tasks (image classification, human pose estimation and semantic segmentation) under extremely low FLOPs. We hope this work provides good baselines for efficient CNNs on multiple vision tasks.

\section{Appendix}
In the appendix, we show the details of MicroNet architectures for human pose estimation. 

\subsection{Architectures for Human Pose Estimation} \label{apx:arch-kp}

We use COCO 2017 dataset \cite{lin2014microsoft} to evaluate MicroNet on single-person keypoint detection. The human detection boxes are cropped and resized to $256\times192$, and each person instance is labeled with 17 keypoints. 

\begin{table*}[b!]
	\begin{center}
	    \footnotesize
	    \setlength{\tabcolsep}{0.5mm}{
		\begin{tabular}{@{\hskip 1mm}c|c@{\hskip 2.2mm}c@{\hskip 2.2mm}c@{\hskip 2.2mm}c@{\hskip 2.2mm}c|c@{\hskip 2.2mm}c@{\hskip 2.2mm}c@{\hskip 2.2mm}c@{\hskip 2.2mm}c|c@{\hskip 2.2mm}c@{\hskip 2.2mm}c@{\hskip 2.2mm}c@{\hskip 2.2mm}c|c@{\hskip 2.2mm}c@{\hskip 2.2mm}c@{\hskip 2.2mm}c@{\hskip 2.2mm}c@{\hskip 1mm}}
		    \specialrule{.1em}{.05em}{.05em} 
			& \multicolumn{5}{c|}{M0} & \multicolumn{5}{c|}{M1} & \multicolumn{5}{c|}{M2} & \multicolumn{5}{c}{M3}   \\
			\cline{2-21}
			Output & Block & $k$ & $C$ & $\frac{C}{R}$ & $G$ & Block & $k$ & $C$ & $\frac{C}{R}$ & $G$ & Block & $k$ & $C$ & $\frac{C}{R}$ & $G$ & Block & $k$ & $C$ & $\frac{C}{R}$ & $G$\\
		
			\specialrule{.1em}{.05em}{.05em} 
			backbone &  &  &  &  &  &   &  &  &  &  &   &  & &  &  &   &  &  & &  \\
			\hline
			128$\times$96 &  stem & 3 & 12 & 4 & (1, 4) &  stem & 3 & 12 & 4 & (1, 4) &  stem & 3 & 16 & 4 & (1, 4) &  stem & 3 & 16 & 4 & (1, 4)\\
			&  Micro-A & 3 & 12 & 12 & (3, --) &  Micro-A & 3 & 12 & 12 & (3, --) &  Micro-A & 3 & 16 & 16 & (4, --) &  Micro-A & 3 & 16 & 16 & (4, --)  \\
			\hline
			64$\times$48 &  Micro-A & 3 & 48 & 16 & (4, --) &  Micro-A & 3 & 48 & 16 & (4, --) &  Micro-A & 3 & 64 & 24 & (4, --) &  Micro-A & 3 & 64 & 24 & (4, --)  \\
			&  Micro-A & 5 & 64 & 24 & (4, --) &  Micro-A & 5 & 64 & 24 & (4, --) &  Micro-B & 5 & 192 & 32 & (4, 8) &  Micro-B & 5 & 128 & 32 & (4, 8)  \\
			&  &  &  &  &  &   &  &  &  &  &  & &  &  &  &  Micro-C & 3 & 192 & 48 & (8, 6)  \\
			\hline
			&  Micro-B & 5 & 192 & 32 & (4, 8) &  Micro-B & 5 & 192 & 32 & (4, 8) &  Micro-C & 5 & 288 & 48 & (6, 8) &  Micro-C & 5 & 256 & 64 & (8, 8)  \\
			32$\times$24 & Micro-C & 7 & 288 & 48 & (6, 8) & Micro-C & 7 & 288 & 48 & (6, 8) & Micro-C & 7 & 384 & 64 & (8, 8) &  Micro-C & 7 & 256 & 64 & (8, 8)  \\
			&   &  &  &  &  &  Micro-C & 5 & 384 & 64 & (8, 8) & Micro-C & 5 & 480 & 80 & (8, 10) &  Micro-C & 7 & 256 & 64 & (8, 8)  \\
			&  &  &  &  &  &   &  &  &  &  &  & &  &  &  &  Micro-C & 5 & 384 & 96 & (8, 12)  \\
			&  &  &  &  &  &   &  &  &  &  &  & &  &  &  &  Micro-C & 5 & 384 & 96 & (8, 12)  \\
			\hline
			& Micro-C & 5 & 384 & 64 & (8, 8) & Micro-C  & 5 & 576 & 96 & (8, 12) &  Micro-C & 5 & 720 & 120 & (10, 12) &  Micro-C & 5 & 576 & 144 & (12, 12)  \\
			& Micro-C & 7 & 576 & 96 & (8, 12) & Micro-C  & 7 & 768 & 128 & (8, 16) &  Micro-C & 7 & 720 & 120 & (10, 12) &  Micro-C & 5 & 576 & 144 & (12, 12)  \\
			16$\times$12&  &  &  &  &  &  Micro-C & 5 & 768 & 128 & (8, 16) & Micro-C & 5 & 720 & 120 & (10, 12) &  Micro-C & 5 & 576 & 144 & (12, 12)  \\
			&  &  &  &  &  &   &  &  &  &  &  Micro-C & 7 & 864 & 144 & (12, 12) &  Micro-C & 7 & 768 & 192 & (12, 16)  \\
			&  &  &  &  &  &   &  &  &  &  &   &  & &  &  &  Micro-C & 5 & 768 & 192 & (12, 16)  \\
			&  &  &  &  &  &   &  &  &  &  &   &  & &  &  &  Micro-C & 5 & 1024 & 256 & (16, 16)  \\
			\specialrule{.1em}{.05em}{.05em} 
			head &  &  &  &  &  &   &  &  &  &  &   &  & &  &  &   &  &  & &  \\
			\hline
			32$\times$24&  Micro-C & 5 & 480 & 120 & (12, 10) &  Micro-C & 5 & 640 & 160 & (16, 10) & Micro-C  & 5 & 768 & 192 & (12, 16) &  Micro-C & 5 & 1024 & 256 & (16, 16)  \\
			\hline
			64$\times$48&  Micro-C & 5 & 320 & 80 & (8, 10) &  Micro-C & 7 & 384 & 96 & (8, 12) & Micro-C  & 7 & 448 & 112 & (8, 14) &  Micro-C & 7 & 640 & 160 & (16, 10)  \\
			&  Micro-A$^\dag$ & 5 & 320 & 48 & (8, --) &  Micro-A$^\dag$ & 5 & 384 & 64 & (8, --) & Micro-A$^\dag$  & 5 & 448 & 80 & (8, --) &  Micro-A$^\dag$ & 7 & 640 & 96 & (8, --)  \\
			\specialrule{.1em}{.05em}{.05em} 
			 &  \multicolumn{5}{c|}{77.7M MAdds, 1.0M Param} & \multicolumn{5}{c|}{116.8M MAdds, 1.8M Param} & \multicolumn{5}{c|}{163.2M MAdds, 2.2M Param} & \multicolumn{5}{c}{263.2M MAdds, 4.0M Param} \\
			\specialrule{.1em}{.05em}{.05em} 
		\end{tabular}
		}
	\end{center}
	\vspace{-1mm}
	\caption{\textbf{MicroNet Architectures for Keypoint Detection}. ``stem" refers to stem layer. ``Micro-A", ``Micro-B", and ``Micro-C" refers to three Micro-Blocks (see section \ref{section:micro-arch} and Figure \ref{fig:micro-block} for more details). $k$ is the kernel size, $C$ is the number of output channels, $R$ is the channel reduction ratio in Micro-Factorized pointwise convolution, $G$ is the group number. Note that for ``Micro-A" that uses the lite combination of Micro-Factorized pointwise and depthwise convolutions, $C$ is the number of output channels in Micro-Factorized depthwise convolution, $\frac{C}{R}$ is the number of output channels for the block. The last block in the head is Micro-A$^\dag$, in which Micro-Factorized depthwise convolution does \textit{not} expand the number of channels. }
	\vspace{-1mm}
	\label{table:micro-arch-kp}
\end{table*}

Table \ref{table:micro-arch-kp} shows the four MicroNet models (M0--M3) for human pose estimation. Different from ImageNet classification, the backbone for keypoint detection stops at the stride of 16 (resolution $16 \times 12$) to retain more spatial information. Each model has a head that includes three Micro-Blocks. The first one is with stride of 8 (resolution $32 \times 24$), while the other two are with stride of 4 (resolution $64 \times 48$). Note that the last block is Micro-Block-A, which uses the lite combination to shrink the number of channels. It is different from Micro-Block-A in the backbone, as its Micro-Factorized depthwise convolution does \textit{not} expand the number of channels. The head is then followed by a pointwise convolution to generate heatmaps for 17 keypoints. We use bilinear upsampling to increase resolution in the head, and use the spatial attention in \cite{Chen2020DynamicReLU} per layer.


{\small
\bibliographystyle{ieee_fullname}
\bibliography{egbib}
}

\end{document}